\documentclass[fleqn,10pt]{wlscirep}
\usepackage[utf8]{inputenc}
\usepackage{todonotes}
\usepackage{booktabs}
\usepackage{longtable}
\usepackage{graphicx}
\usepackage{subcaption}
\usepackage{multirow}
\usepackage{comment}
\usepackage{pifont}
\usepackage{hyperref}
\newcommand{\cmark}{\textcolor{green!70!black}{\ding{51}}}
\newcommand{\xmark}{\textcolor{red}{\ding{55}}}
\usepackage[T1]{fontenc}
\title{A Comprehensive Review of Datasets for Clinical Mental Health AI Systems}

\author[1,2]{Aishik Mandal}
\author[3,4]{Prottay Kumar Adhikary}
\author[1]{Hiba Arnaout}
\author[1]{Iryna Gurevych}
\author[3,4,*]{Tanmoy Chakraborty}
\affil[1]{Ubiquitous Knowledge Processing Lab (UKP Lab)\\
Department of Computer Science and Hessian Center for AI (hessian.AI)\\
Technische Universität Darmstadt}
\affil[2]{National Research Center for Applied Cybersecurity ATHENE, Germany}
\affil[3]{Department of Electrical Engineering, Indian Institute of Technology Delhi, India}
\affil[4]{Yardi School of Artificial Intelligence, Indian Institute of Technology Delhi, India}

 \affil[*]{Corresponding author: Tanmoy Chakraborty (tanchak@iitd.in)}

\begin{abstract}
Mental health disorders are rising worldwide. However, the availability of trained clinicians has not scaled proportionally, leaving many people without adequate or timely support. To bridge this gap, recent studies have shown the promise of Artificial Intelligence (AI) to assist mental health diagnosis, monitoring, and intervention. However, the development of efficient, reliable, and ethical AI to assist clinicians is heavily dependent on high-quality clinical training datasets. Despite growing interest in data curation for training clinical AI assistants, existing datasets largely remain scattered, under-documented, and often inaccessible,  hindering the reproducibility, comparability, and generalizability of AI models developed for clinical mental health care. In this paper, we present the first comprehensive review of clinical mental health datasets relevant to the training and development of AI-powered clinical assistants. We categorize these datasets by mental disorders (e.g., depression, schizophrenia), data modalities (e.g., text, speech, physiological signals), task types (e.g., diagnosis prediction, symptom severity estimation, intervention generation), accessibility (public, restricted or private), and sociocultural context (e.g., language and cultural background). Along with these, we also investigate synthetic clinical mental health datasets. We identify critical gaps such as a lack of longitudinal data, limited cultural and linguistic representation, inconsistent collection and annotation standards, and a lack of modalities in synthetic data. We conclude by outlining key challenges in curating and standardizing future datasets and provide actionable recommendations to facilitate the development of more robust, generalizable, and equitable mental health AI systems.

\if 
Increasing mental health issues and a lack of qualified clinical experts mean a growing number of people are going untreated. This has resulted in growing interest in using Artificial Intelligence to assist clinicians and enable them to treat more patients. Training such mental health AI assistants necessitates the creation of dedicated clinical mental health datasets. While there are many works curating datasets for various mental health issues, there is a lack of a comprehensive review of clinical mental health datasets. In this work, we review datasets from different perspectives: types of mental health issues, accessibility, types of tasks, modalities, and language/ cultural background. This paper will help future mental health applications determine which dataset is suitable for training and evaluating their AI systems. Moreover, we will highlight key issues with current datasets, discuss challenges towards creating better datasets, and suggest possible directions to create better generalized datasets.

\fi

\end{abstract}
\begin{document}

\flushbottom
\maketitle
%Click the title above to edit the author information and abstract
%
\thispagestyle{empty}

% \noindent Please note: Abbreviations should be introduced at the first mention in the main text – no abbreviations lists. Suggested structure of main text (not enforced) is provided below.

\section*{Introduction}

Mental health disorders are a growing global concern, affecting millions and placing increasing pressure on already strained healthcare systems. Effective treatment typically involves trained clinicians conducting assessments, developing individualized care plans, and engaging in ongoing therapeutic sessions. However, such clinician-centered care is time-intensive, and a persistent global shortage of mental health professionals leaves many individuals without adequate or timely support. To address this care gap, there is growing interest in applying artificial intelligence (AI) to augment mental health diagnosis, monitoring, and intervention. AI systems offer the potential to improve clinical decision-making, scale access to care, and personalize treatment. However, the development of reliable and generalizable mental health AI tools critically depends on access to high-quality training and evaluation datasets.

The collection and dissemination of clinical mental health data is hampered by stringent privacy regulations, such as the General Data Protection Regulation (GDPR) \cite{gdpr} in the European Union and the Health Insurance Portability and Accountability Act (HIPAA) \cite{act1996health} in the United States. These frameworks place strict limitations on the use and sharing of sensitive health information, complicating the creation of large, representative datasets that meet ethical and legal standards. Early data-driven approaches often turned to social media platforms such as Reddit~\cite{reddit-1}, Twitter~\cite{twitter-1}, and YouTube~\cite{yt-1} to compile mental health datasets~\cite{guntuku2017detecting}. While these sources offer scale and accessibility, they generally lack clinical validation and may not accurately reflect diagnosed populations, limiting their clinical applicability.

In recent years, researchers have begun constructing clinically grounded datasets through collaborations with mental health institutions, often under strict privacy-preserving protocols and with informed consent. These datasets, derived from electronic health records, clinician notes, therapy transcripts, and physiological or neuroimaging data, are often shared \textit{only} upon request or remain private, following extensive pseudonymization procedures.  Despite increased interest in such resources, there is no comprehensive review focused specifically on clinical mental health datasets for AI development. Previous reviews have addressed broader applications of AI in mental health~\cite{ai_review_2}, or specific methodological areas such as machine learning~\cite{ml_review_1}, deep learning~\cite{mldl_review_1}, natural language processing (NLP)\cite{nlp_review_1}, and large language models (LLMs)\cite{llm_review_1}.

\begin{figure}[!t]
\centering
\includegraphics[width=0.7\linewidth]{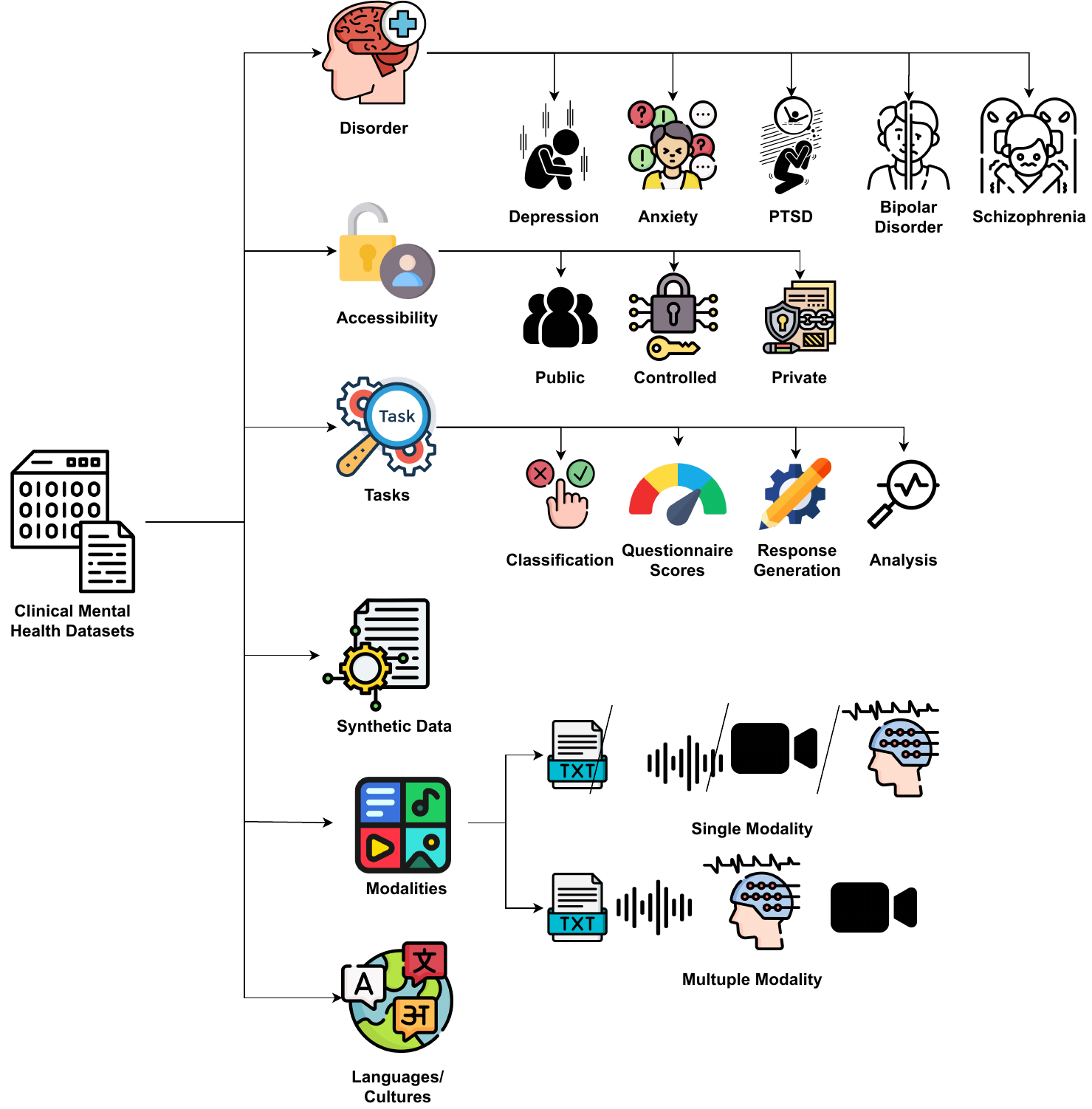}
\caption{Schematic overview of this review. This diagram illustrates the key dimensions covered in our review, including mental health disorder categories, dataset accessibility levels, task types, synthetic data sources, data modalities, and cultural or linguistic representation. The taxonomy offers a structured framework for understanding the landscape of clinical mental health datasets and serves as a guide for navigating available resources in this domain.}
\label{fig:schematic}
\end{figure}

Although some recent review articles (see Table~\ref{tab:novelty}) have incorporated data-centric perspectives, they often treat datasets as secondary to model performance or application domains. Reviews have examined resources spanning electronic health records, imaging, social media, wearables, and digital platforms~\cite{graham2019artificial,dehbozorgi2025application,shatte2019machine,su2020deep,healthcare11030285}, or focused on natural language processing~\cite{scherbakov2025natural,laricheva2024scoping,malgaroli2023natural}, yet few provide a systematic analysis of dataset characteristics. Key considerations, such as accessibility (public, restricted, private), cultural and linguistic representation, and the emerging role of synthetic data, remain largely unaddressed. In contrast, our review provides a dedicated and comprehensive examination of clinical mental health datasets for AI development. We categorize datasets by disorder type, access level, task formulation, data modality, and sociocultural context, then identify gaps that limit model generalizability and clinical relevance. We further highlight current efforts in synthetic data generation to address privacy and data scarcity. By shifting the focus from algorithms to data foundations, our work offers actionable guidance for curating and evaluating datasets that support robust, ethical, and equitable mental health AI systems (see Figure~\ref{fig:schematic}).

To identify relevant resources for this survey, we conducted a systematic search across Google Scholar, DBLP, and the ACL Anthology, using a set of targeted keywords covering both general and disorder-specific domains. These included terms such as ``Mental Health AI Datasets,'' ``Multimodal Mental Health AI Datasets,'' ``Mental Health AI Datasets for Depression,'' ``Multimodal Mental Health AI Datasets for Depression,'' ``Mental Health AI Datasets for Anxiety,'' ``Multimodal Mental Health AI Datasets for Anxiety,'' ``Mental Health AI Datasets for Bipolar Disorder,'' ``Multimodal Mental Health AI Datasets for Bipolar Disorder,'' ``Mental Health AI Datasets for PTSD,'' ``Multimodal Mental Health AI Datasets for PTSD,'' ``Mental Health AI Datasets for Schizophrenia,'' ``Multimodal Mental Health AI Datasets for Schizophrenia,'' ``MRI Mental Health AI Datasets,'' and ``EEG Mental Health AI Datasets''. This search initially yielded $560$ records, which, after deduplication and screening for relevance, were narrowed down to $89$ datasets that met our selection criteria of containing real-world clinical mental health data applicable to AI research.

\begin{table}[!t]
\centering
\begin{tabular}{@{}lllccccccccc@{}}
\toprule
\multirow{2}{*}{Review} & \multirow{2}{*}{Focus}& \multirow{2}{*}{Data Source} & \multirow{2}{*}{\#Datasets} & \multicolumn{5}{c}{Aspects covered}& \multirow{2}{*}{\begin{tabular}[c]{@{}c@{}}Synthetic \\ Data\end{tabular}} \\ 
\cmidrule(lr){5-9}
 & & & & D & A & T & M & C &\\ 
\midrule

Garg et al. \cite{social_media_review_1} & {\begin{tabular}[c]{@{}l@{}}Algorithms, \\ Datasets\end{tabular}} & SM &17 &\cmark&\cmark&\xmark&\xmark& \xmark & \xmark \\

Ahmed et al. \cite{data_review_2} & Datasets &Clinical, SM &23 &\cmark&\cmark&\cmark&\xmark& \cmark & \xmark \\
Thieme et al. \cite{hci} & HCI &Clinical, SM &NA &\cmark&\xmark&\cmark&\cmark& \xmark & \xmark \\
El-Sappagh et al. \cite{trustworthy}& Responsible AI &Clinical &49 &\cmark&\cmark&\cmark&\cmark& \xmark & \xmark \\
Uyanik et al. \cite{neuro1} & Multimodal Algorithms &Clinical &15 &\cmark&\xmark&\xmark&\xmark& \xmark & \xmark \\
Tyagi et al. \cite{neuro2} & Multimodal Algorithms &Clinical &8 &\xmark&\xmark&\xmark&\xmark& \xmark & \xmark \\
Al Sahili et al. \cite{mult1} & {\begin{tabular}[c]{@{}l@{}}Multimodal Algorithms, \\ Datasets\end{tabular}} &YT, Movie &26 &\cmark&\xmark&\xmark&\cmark& \cmark & \xmark \\
Graham et al. \cite{graham2019artificial} & AI Algorithms &EHR, SM & 28 & \cmark & \xmark & \cmark & \cmark & \xmark & \xmark \\
Dehbozorgi et al. \cite{dehbozorgi2025application} & AI Algorithms &Apps, Chatbots & 15 & \cmark & \xmark & \cmark & \cmark & \xmark & \xmark \\
Shatte et al. \cite{shatte2019machine} & ML Algorithms &EHR, SM & 300 & \cmark & \xmark & \cmark & \cmark & \xmark & \xmark \\
Su et al. \cite{su2020deep} & DL Algorithms &Clinical, SM & 57 & \cmark & \xmark & \cmark & \cmark & \xmark & \xmark \\
Iyortsuun et al. \cite{healthcare11030285} & ML/DL Algorithms &EHR, SM 
  & 33 & \cmark & \cmark & \cmark & \cmark & \xmark & \xmark \\
Scherbakov et al. \cite{scherbakov2025natural} & NLP Algorithms &Notes, SM & 1768 & \cmark & \cmark & \cmark & \cmark & \xmark & \xmark \\
Laricheva et al. \cite{laricheva2024scoping} & NLP Algorithms &Transcripts & 41 & \cmark & \xmark & \cmark & \cmark & \xmark & \xmark \\
Malgaroli et al. \cite{malgaroli2023natural} & NLP Algorithms &Clinical, Apps & 102 & \cmark & \cmark & \cmark & \cmark & \xmark & \xmark \\
\midrule
\textbf{Ours} & Datasets &Clinical & 89  & \cmark & \cmark & \cmark & \cmark & \cmark & \cmark \\ 
\bottomrule
\end{tabular}
\caption{Compared to prior reviews, our review uniquely focuses on clinical mental health datasets, offering a systematic analysis by disorder type, data accessibility, task type, modality, and cultural context. We are also the first to review synthetic data generation for clinical mental health applications. Here D = Mental Disorders, A = Access Level, T = Task type, M = Modalities and C = Cultural background. SM represent Social Media and EHR denotes Electronic Health Record.}
\label{tab:novelty}
\end{table}

\begin{longtable}[!t]{@{}llllllll@{}}
\toprule
Dataset & Modalities & Disorder& Task& Size (P) & Imb (Dis./HC) & Culture& Access \\* \midrule
\endfirsthead
\endhead
\bottomrule
\endfoot
\endlastfoot
%

% Tang et al. \cite{t_1} & T & SZ& BC & 31 (31) &20/11 & US& Restricted \\
LabWriting \cite{t_2} & T & SZ& BC & 373 (188)& 93/95& US& Private \\
% Wawer et al. \cite{t_3}& T & SZ& BC & 94 (94)& 47/47& Poland& Private \\
% Hong et al. \cite{t_4} & T & SZ& BC & 201 (39)&23/16 & US& Private \\
% Allende-Cid et al. \cite{t_5} & T & SZ& BC & 189 (63) & 13/50&Chile & Private \\
% Iter et al. \cite{t_6} & T & SZ& BC & 14 (14)&9/5 & US& Private \\
% Elvevåg et al. \cite{t_7} & T & SZ& Analysis & 83 (83) & 53/30& US& Private \\
Gehrmann et al. \cite{gehrmann2018comparing} & T & \begin{tabular}[c]{@{}l@{}}Dep. \&\\ SZ\end{tabular}&BC & \begin{tabular}[c]{@{}l@{}}41,000 \\ (41,000)\end{tabular} & NA & US & Restricted \\
WorryWords \cite{mohammad-2024-worrywords} & T & Anx. & BC & \begin{tabular}[c]{@{}l@{}}44,450 \\ (44,450)\end{tabular} & NA & Canada & Public\\
Hou et al. \cite{t_108} & T & Dep. & Analysis & 376 (376) & 128/248 & China & Private \\
CHSN \cite{t_107} & T & \begin{tabular}[c]{@{}l@{}}Dep. \&\\ Anx. \&\\ BD\\\end{tabular} & \begin{tabular}[c]{@{}l@{}}BC,\\ Analysis\end{tabular} & 6M (1M) & NA & US & Private \\
PRIORI \cite{a_1} & A & \begin{tabular}[c]{@{}l@{}}Dep. \&\\ BD\end{tabular}& MC & 34,830 (37) &NA & US& Private \\
FME Hospital Dataset \cite{a_2} & A & PTSD& MC & 200 (200) & 150/50 & Taiwan & Public \\
PTSD Speech Corpus \cite{a_3} & A & PTSD& BC & 26 (26)&8/18 & US& Private \\
German LMU Dataset \cite{a_4} & A& PTSD& BC & 15 (15) &7/8 & Germany& Private \\
Vergyri et al. \cite{a_5} & A & PTSD& BC & 39 (39) &15/24 & US& Private \\
Marmar et al. \cite{a_6} & A & PTSD & BC & 129 (129) &52/77 & US& Private \\
Hu et al. \cite{a_7} & A & PTSD& BC/QS & 136 (136)& 76/60& China& Private \\
RADAR-MDD \cite{t_110} & A & Dep. & Analysis & 585 (585) & NA & \begin{tabular}[c]{@{}l@{}}UK/ \\ Spain/ \\ Netherlands\end{tabular} & Restricted \\
Chang et al. \cite{t_106} & A & Dep. & Analysis & 62 (25) & 25/0 & US & Private \\
Broek et al. \cite{t_105} & A & PTSD & MC & 24 (24) & 24/0 & Netherlands & Private \\
Salekin et al. \cite{t_104} & A & \begin{tabular}[c]{@{}l@{}}Dep. \&\\ Anx.\end{tabular} & BC & 105 (105) & 60/45 & US & Private \\
Abbas et al. \cite{v_1} & V & SZ& BC/QS & 27 (27)&18/9 & US& Private \\
Shafique et al. \cite{v_2} & V & Anx.& MC & 50 (50) &40/10 & Pakistan& Public \\
Langer et al. \cite{v_3} & V & Anx.& Analysis & 114 (114)&65/49 & US& Private \\
Pampouchidou et al. \cite{v_4} & V & \begin{tabular}[c]{@{}l@{}}Dep. \&\\ Anx.\end{tabular}& BC/QS & 322 (65) &20/45 & Greece& Private \\
Jiang et al.\cite{v_6}& V& Dep.& BC & 365 (12) & NA &US & Private\\
Gilanie et al.\cite{v_5}& V& BD& BC & 502 (502)& 310/192 & Pakistan& Private\\
EATD \cite{ta_1} & T, A & Dep. & BC & 162 (162)&30/132 & China& Public \\
Aloshban et al.\cite{ta_2} & T, A & Dep.& BC & 59 (59) &29/30& Italy& Private\\
MMPsy \cite{ta_3} & T, A & \begin{tabular}[c]{@{}l@{}}Dep.\&\\ Anx.\end{tabular} & BC & \begin{tabular}[c]{@{}l@{}}11,983 \\ (11,983) \end{tabular} & 1,557/10,426& China & Public \\
Tang et al. \cite{t_1} & T, A & SZ& BC & 31 (31) &20/11 & US& Restricted \\
Wawer et al. \cite{t_3}& T, A & SZ& BC & 94 (94)& 47/47& Poland& Private \\
Hong et al. \cite{t_4} & T, A & SZ& BC & 201 (39)&23/16 & US& Private \\
Allende-Cid et al. \cite{t_5} & T, A & SZ& BC & 189 (63) & 13/50&Chile & Private \\
Iter et al. \cite{t_6} & T, A & SZ& BC & 14 (14)&9/5 & US& Private \\
Elvevåg et al. \cite{t_7} & T, A & SZ& Analysis & 83 (83) & 53/30& US& Private \\
Bedi et al.\cite{ta_4} & T, A & SZ& BC & 34 (34)&5/29& US& Private\\
Elvevag et al.\cite{ta_5}& T, A & SZ & Analysis& 51 (51)& 26/25 & US & Private\\
Li et al. \cite{ta_6} & T, A & SZ& BC/QS & 63 (63)& 38/25 & China & Private\\
Ciampelli et al. \cite{ta_7} & T, A & SZ& BC/QS & 163 (163)&93/70 & Netherlands & Private\\
Cabuk et al. \cite{ta_8} & T, A & SZ & Analysis & 76 (76)& 38/38& Turkey & Private\\
Xu et al. \cite{ta_9}& T, A & SZ& BC/QS& 75 (75) &50/25& Singapore & Private\\
Jeong et al. \cite{ta_10} & T, A & SZ& Analysis& 22 (7) & 7/0 & Canada & Private\\
% Holmlund et al. \cite{ta_11}& T, A & Schizophrenia & Binary & 104 && US & Restricted \\
Parola et al. \cite{ta_12} & T, A & SZ& Analysis & 387 (387)&187/200& \begin{tabular}[c]{@{}l@{}}Denmark/\\ China/\\ Germany\end{tabular} &Private\\
Aich et al.\cite{ta_13} & T, A & \begin{tabular}[c]{@{}l@{}}SZ \&\\ BD\end{tabular} & MC & 1288 (644)& \begin{tabular}[c]{@{}l@{}}247 SZ\\286 BD \\ 110 HC \end{tabular} & US & Public \\
Arslan et al. \cite{ta_14}& T, A & BD& MC & 143 (143) & \begin{tabular}[c]{@{}l@{}}53 FEP\\40 FEBD \\50HC\end{tabular} & Turkey&Private\\
DEPAC \cite{t_111} & T, A & \begin{tabular}[c]{@{}l@{}}Dep. \&\\ Anx.\end{tabular} & Analysis & 2674 (571) & NA & Canada & Private \\
Ex-ray \cite{t_103} & T, A & SZ & BC & 56 (56) & 47/9 & Australia & Private \\
Hayati et al. \cite{t_101} & T, A & Dep. & BC & 53 (53) & 11/42 & Malaysia & Private \\
Jiang et al.\cite{av_2} & A, V & \begin{tabular}[c]{@{}l@{}}Dep. \&\\ Anx.\end{tabular}& BC & 73 (73)&51/22& US& Private\\
AViD Corpus\cite{av_4} & A, V & Dep.& QS& 340 (292)& NA &Germany& Restricted \\
Pittsburgh\cite{av_5} & A, V & Dep.& MC& 130 (49)&49/0&US& Restricted \\
Black Dog Institute\cite{av_6} & A, V & Dep.&BC & 60 (60)&30/30& Australia& Private \\
Lin et al.\cite{av_7} & A, V & \begin{tabular}[c]{@{}l@{}}Dep. \&\\ Anx.\end{tabular}& BC/QS &35 (35)& 18/17 & UK& Restricted\\
Guo et al.\cite{av_8} & A, V & Dep.& BC& 208 (208)&104/104& China& Restricted\\
E-DAIC\cite{edaic}& T, A, V& \begin{tabular}[c]{@{}l@{}}Dep.\&\\ PTSD\end{tabular}& QS & 275 (275) &66/209& US & Restricted \\
CMDC\cite{tav_1}& T, A, V& Dep.& BC/QS &78 (78) &26/52& China& Public \\
VH DAIC \cite{tav_2}& T, A, V& \begin{tabular}[c]{@{}l@{}}Dep. \&\\ PTSD\end{tabular} & BC & 53 (53) &\begin{tabular}[c]{@{}l@{}}22/31(PTSD) \\\& 17/36(Dep.)\end{tabular} & US& Private\\
Schultebraucks et al. \cite{tav_3} & T, A, V& \begin{tabular}[c]{@{}l@{}}Dep. \&\\ PTSD\end{tabular}& BC & 81 (81)&NA& US& Private\\
MEDIC\cite{tav_4} & T, A, V& Coun.& BC &38 (10)& NA & China & Restricted \\
BDS\cite{tav_5} & T, A, V& BD&BC/MC& 95 (95)&46/49& Turkey & Restricted \\
Chuang et al. \cite{tav_6}& T, A, V& SZ& QS& 37 (26) &26/0& Taiwan & Private\\
Premananth et al. \cite{tav_7}& T, A, V& SZ& MC & 140 (40)&30/16 & US & Private\\
Zhang et al. \cite{tav_8} & T, A, V& SZ& BC & 160 (40) & 20/20 & China& Restricted \\
Tao et al. \cite{t_102} & T, A, V & \begin{tabular}[c]{@{}l@{}}Dep. \&\\ Anx.\end{tabular} & BC & 139 (139) & 64/75 & China & Private \\
% Depresjon \cite{Garcia:2018:NBP:3083187.3083216} & Actigraphy & Depression & BC & 55 & NA & Norway & Public\\
MODMA\cite{modma} & A, EEG & Dep. &QS & 52 (52) &23/29&China & Restricted \\
VerBIO \cite{9311251} & A, EEG & Anx. & Analysis & 344 (344) & NA& US & Restricted\\
COBRE \cite{cobre}& sMRI, fMRI & SZ & BC &146 (146)& 72/74 &US& Public \\
Park et al. \cite{t_137} & EEG & \begin{tabular}[c]{@{}l@{}}Dep. \&\\ Anx. \&\\ SZ \end{tabular} & BC & 945 (945) & NA & South Korea & Public \\
RepOD \cite{t_136} & EEG & SZ & Analysis & 28 (28) & 14/14 & Poland & Public \\
SchizConnect \cite{t_135} & EEG & SZ & MC & 1392 (1392) & NA & US & Public \\
UCLA \cite{t_134} & fMRI, sMRI & \begin{tabular}[c]{@{}l@{}} SZ \&\\ BD \end{tabular} & MC & 229 (229) & \begin{tabular}[c]{@{}l@{}}50 SZ\\ 49 BD \\ 130 HC\end{tabular} & US & Restricted \\
NUSDAST \cite{t_133} & MRI & SZ & BC & 341 (341) & 171/170 & US & Public \\
MCIC \cite{t_132} & fMRI, sMRI & SZ & \begin{tabular}[c]{@{}l@{}}BC,\\ Analysis\end{tabular} & 331 (331) & 162/169 & US & Public \\
MLSP2014 \cite{t_131} & fMRI, sMRI & SZ & BC & 144 (144) & 69/75 & US & Public \\
FBIRN \cite{t_130} & fMRI, sMRI & SZ & BC & 256 (256) & 128/128 & US & Public \\
Cai et al. \cite{t_129} & EEG & Dep. & BC & 213 (213) & 92/121 & China & Private \\
Pend et al. \cite{t_128} & EEG & Dep. & BC & 55 (55) & 27/28 & China & Private \\
Zhu et al. \cite{t_127} & A, V, EEG & Dep. & BC & 51 (51) & 24/27 & China & Private \\
Mumtaz et al. \cite{t_126} & EEG & Dep. & BC & 64 (64) & 34/30 & Malaysia & Public \\
Cavanagh et al. \cite{t_124} & EEG & Dep. & BC & 122 (122) & 46/76 & US & Public \\
Luo et al. \cite{t_123} & EEG & Dep. & BC & 40 (40) & 18/22 & China & Private \\
Garg et al. \cite{t_122} & EEG & Dep. & BC & 120 (120) & 62/58 & Malaysia & Public \\
Li et al. \cite{t_121} & EEG & Dep. & BC & 140 (140) & 70/70 & China & Restricted \\
Shen et al. \cite{t_120} & EEG & Dep. & BC & \begin{tabular}[c]{@{}l@{}}35 (35)\\ 170 (170)\\ 214(214)\end{tabular} & \begin{tabular}[c]{@{}l@{}}15/20 \\ 81/89 \\ 105/109\end{tabular} & China & Private \\
Chung et al. \cite{t_119} & EEG & Dep. & BC & 214 (67) & 49/18 & Taiwan & Restricted \\
PRED + CT \cite{t_118} & EEG & Dep. & BC & 119 (119) & 44/75 & US & Public \\
Ros et al. \cite{t_117} & EEG & PTSD & BC & 50 (50) & 20/30 & Canada & Private \\
Nicholson et al. \cite{t_116} & EEG & PTSD & BC & 73 (73) & 41/32 & Canada & Private \\
Kim et al. \cite{t_115} & EEG & SZ & BC & 238 (238) & 119/119 & South Korea & Private \\
Barros et al. \cite{t_114} & EEG & SZ & BC & 128 (128) & 65/63 & US & Public \\
Borisov et al. \cite{t_113} & EEG & SZ & BC & 84 (84) & 45/39 & Russia & Public \\
MPRC \cite{t_112} & EEG & SZ & BC & 78 (78) & 46/32 & US & Public \\
SRPBS \cite{t_109} & MRI & \begin{tabular}[c]{@{}l@{}}Dep. \&\\ SZ\end{tabular}& Analysis & 2030 (2030) & \begin{tabular}[c]{@{}l@{}}450 BD\\ 159 SZ \\1421 HC\end{tabular} & Japan & Restricted \\

\bottomrule
% ADHD 200& SocialsRI, fMRI & ADHD& Binary Classification & && &\\* \bottomrule
\caption{We provide a detailed summary of clinical mental health datasets, outlining key characteristics: modality (T: text, A: audio, V: video), associated disorder (SZ: schizophrenia, Dep.: depression, Anx.: anxiety, BD: bipolar disorder, PTSD: post-traumatic stress disorder), task type (BC: binary classification, MC: multi-class classification, QS: questionnaire score prediction, Reg.: regression), dataset size and number of participants (P), data imbalance (Imb.; Dis/HC = participants with disorder / healthy controls, FEP: first episode psychosis, FEBD: first episode bipolar disorder), cultural context, and accessibility level. Coun. stands for counseling.}
\label{tab:real-datasets}
\end{longtable}

\section*{Mental Disorders}

\begin{figure}[!t]
\centering
\includegraphics[width=\linewidth]{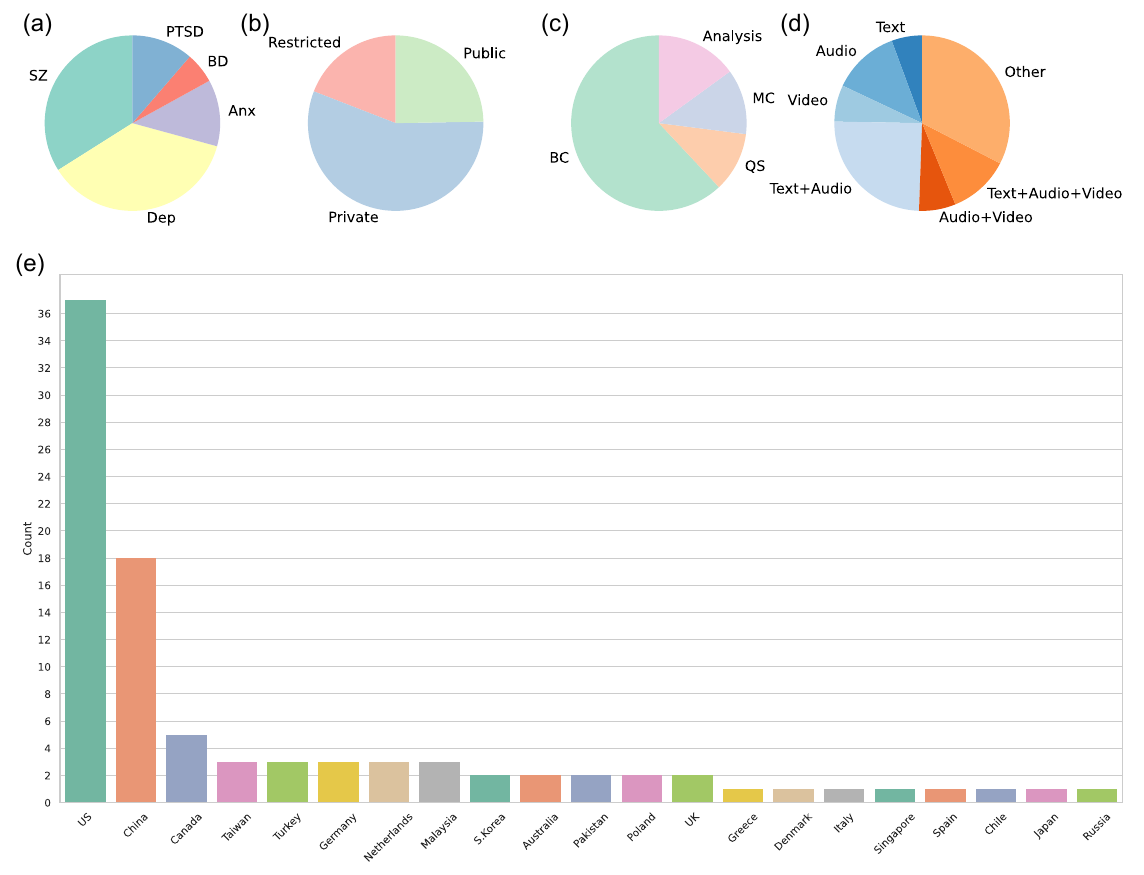}
\caption{An overview of the distribution of clinical mental health datasets reviewed in this work. \textbf{(a)} Distribution by mental health disorder, including schizophrenia (SZ), depression (Dep), anxiety (Anx), bipolar disorder (BD), and post-traumatic stress disorder (PTSD). \textbf{(b)} Distribution by dataset accessibility level: public, restricted, and private. \textbf{(c)} Distribution by task type: binary classification (BC), multi-class classification (MC), questionnaire score prediction (QS), and analysis tasks. \textbf{(d)} Distribution by modality and modality combinations: text, audio, video, text+audio, audio+video, text+audio+video, and others (e.g., EEG, MRI). \textbf{(e)} Distribution showing the number of datasets collected in each country. }
\label{fig:graphs}
\end{figure}

According to the World Health Organization (WHO), mental disorders are clinically significant disturbances in cognition, emotion regulation, or behavior, often resulting in distress or impaired functioning. A 2019 WHO report~\footnote{\href{https://www.who.int/news-room/fact-sheets/detail/mental-disorders}{WHO (2019). Mental Health}} estimated  one in eight people globally experiencing a mental disorder, with depression, anxiety, bipolar disorder, post-traumatic stress disorder (PTSD), and schizophrenia being the most prevalent. A more recent study from the National Institute of Mental Health (NIMH) shows one in five adults in America struggling with mental health \footnote{\href{https://www.nimh.nih.gov/health/statistics/mental-illness}{NIH (2022). Mental Illness}}. Figure~\ref{fig:graphs}(a) illustrates the distribution of these conditions across existing clinical mental health datasets. Most datasets focus on schizophrenia, PTSD, and depression, with relatively fewer addressing anxiety and bipolar disorder. Given the distinct symptom profiles of each disorder, condition-specific features, diagnostic criteria, and validated clinical questionnaires are essential. Table \ref{tab:ques} summarizes the primary psychological questionnaires used for assessing the severity of these conditions.

\paragraph{Depression.} Depression is the most frequently represented disorder in clinical datasets, particularly those incorporating multimodal data. Many datasets focus on identifying depressive symptoms through text, speech, and facial expressions \cite{edaic,tav_1,tav_2,tav_3,t_102}. These datasets typically include severity ratings based on standardized scales such as the Hamilton Depression Rating Scale (HDRS)~\cite{hamilton1960rating}, Patient Health Questionnaire. Depression (PHQ-9)~\cite{kroenke2001phq}, Beck Depression Inventory (BDI-II)~\cite{beck1961inventory}, Self rating depression scale (SDS)~\cite{sds}, Quick Inventory of Depressive Symptomatology Self Report (QIDS-SR)~\cite{qids}, Center for Epidemiologic Studies of Depression Scale (CES-D)~\cite{cesd}, Hospital Anxiety and Depression Scale Depression sub scale (HADS-D) \cite{hads} and Inventory of Depressive Symptomatology – Self Report (IDS-SR) \cite{rush2000inventory} making them valuable for both classification and regression tasks. Some other datasets try to diagnose depression through EEG \cite{t_129,t_128,t_126,t_124,t_123,t_122,t_121,t_120,t_119,t_118} or EEG combined with audio and video \cite{modma,t_127}.

\paragraph{Anxiety.} Clinical datasets targeting anxiety are less common and often smaller in size. Most rely on video recordings to capture visual cues like facial tension, restlessness, and gaze aversion \cite{v_2,v_3,v_4}. These datasets are typically collected in clinical interview or exposure settings, and may include self-reported or clinician-administered scores from instruments such as Liebowitz Social Anxiety Scale (LSAS)~\cite{lsas}, State-Trait Anxiety Inventory (STAI)~\cite{stai}, Generalized Anxiety Disorder (GAD-7)~\cite{gad7}, Hospital Anxiety and Depression Scale Anxiety sub scale (HADS-A) \cite{hads}, Social Interaction Anxiety Scale (SIAS) and Social Phobia Scale (SPS) \cite{sias_sps}.

\paragraph{Bipolar Disorder.} Bipolar disorder is underrepresented in publicly-available clinical datasets. Existing resources often include longitudinal data capturing both depressive and manic episodes using audio, text, and video modalities. Speech datasets focus on prosodic rhythm and vocal markers~\cite{a_1}, while text-based corpora capture linguistic markers of mood shifts~\cite{ta_13,ta_14}. A smaller subset includes annotated video data reflecting affective expression~\cite{v_5}. The Young Mania Rating Scale (YMRS)~\cite{ymrs} is typically used for severity labeling when available.

\paragraph{Post-Traumatic Stress Disorder (PTSD).} PTSD is moderately represented in clinical datasets, with a majority focusing on speech-based markers of trauma-related stress. Audio datasets commonly include features such as vocal tension, pitch variability, and temporal disfluencies~\cite{a_2,a_3,a_4,a_5,a_6,a_7,t_105}. Multimodal datasets incorporating text and video also exist, particularly in clinical interview settings~\cite{edaic,tav_2,tav_3}. Some datasets also use EEG for diagnosing PTSD \cite{t_117,t_116}. Severity annotations often rely on the Subjective Unit of Distress (SUD) \cite{sud}, PTSD Checklist for DSM-5 (PCL-5)~\cite{pcl-5} or its civilian variant, the PTSD Checklist – Civilian Version (PCL-C)~\cite{pcl-c}.

\paragraph{Schizophrenia.} Schizophrenia is among the most comprehensively represented disorders in clinical datasets. Many corpora focus on speech disorganization, including spontaneous speech, narrative recall, or interview responses \cite{t_1,t_2,t_3,t_4,t_5,t_6,t_7,ta_4,ta_5,ta_6,ta_7,ta_8,ta_9,ta_10,ta_12,ta_13}. Several also incorporate visual modalities such as head movement~\cite{v_1,tav_8} and facial affect~\cite{tav_7}. In addition, datasets using EEG \cite{t_137,t_136,t_135,t_115,t_114,t_113,t_112} or MRI ~\cite{cobre,t_134,t_133,t_132,t_131,t_130,t_109} data provide access to neurophysiological correlates of psychosis. Commonly used assessments include Scale for the assessment of Positive symptoms (SAPS)~\cite{saps}, Scale for the assessment of Negative symptoms (SANS)~\cite{sans}, Negative Symptom Assessment (NSA-16)~\cite{nsa}, Positive and Negative Syndrome scale (PANSS)~\cite{panss}, and Assessment of Thought, Language and Communication (TLC)~\cite{tlc}, allowing for fine-grained labeling of symptom dimensions.

\section*{Data Accessibility}

Clinical mental health datasets vary widely in their accessibility, directly shaping their utility for research and the risks they pose to patient privacy. We categorize accessibility into three levels: public, restricted, and private, each occupying a different point along the privacy–usability trade-off. While public datasets maximize accessibility, they pose the greatest risk of sensitive information exposure. Private datasets offer the strongest privacy safeguards but are often inaccessible to the broader research community. Restricted datasets represent a compromise, supporting research use under controlled conditions. The distribution of dataset accessibility in our review is shown in Figure~\ref{fig:graphs}(b).

\paragraph{Public Datasets.} Public datasets are freely available for use in training, evaluation, and analysis. While many social media–based mental health datasets are openly released, clinical datasets are rarely public due to the sensitive nature of the data. Even anonymized text, audio, or video samples carry re-identification risks, especially when multiple modalities are combined. Multimodal datasets are particularly vulnerable to cross-modal privacy breaches, where patterns across data types can inadvertently reveal personal identities. Public release also requires explicit participant consent, which is difficult to obtain given understandable concerns about data misuse. As a result, publicly available clinical mental health datasets, especially those with multimodal data, remain scarce (see Table~\ref{tab:real-datasets}). Only some select multimodal datasets from China like EATD \cite{ta_1}, MMPsy \cite{ta_3} and CMDC \cite{tav_1} are made public after rigorous anonymization. Most of the publicly available datasets contain MRI \cite{cobre,t_133,t_132,t_131,t_130} and EEG \cite{t_137,t_136,t_135,t_126,t_124,t_122,t_118,t_114,t_113,t_112} data. A further concern is the misuse of these datasets by actors lacking clinical or ethical oversight, such as unregulated commercial applications of mental health AI.

\paragraph{Restricted Datasets.} Restricted datasets are available to qualified researchers through formal application processes that often include institutional review, ethical approval, and a clearly defined research purpose. This model enables controlled sharing that balances privacy protection with research utility. Although access procedures can delay use, they help ensure that data are not repurposed for non-clinical or commercial ends. Restricted datasets are particularly common for sensitive modalities such as audio and video. For example, the AViD Corpus \cite{av_4}, Pittsburgh dataset \cite{av_5} and datasets from Lin et al. \cite{av_7} and Guo et al. \cite{av_8} contain audio and video modalities and are restricted. Similarly datasets containing text, audio and video modalities like E-DAIC \cite{edaic}, MEDIC \cite{tav_4}, BDS \cite{tav_5} and dataset from Zhang et al. \cite{tav_8} have restricted access. While most datasets with EEG and MRI data are made public, some of them are kept restricted\cite{t_121,t_119,t_134}.

\paragraph{Private Datasets.} Private datasets are held exclusively by the institutions that collected them and are typically not shared beyond internal researchers or a small number of vetted collaborators. They are stored in secure environments, often under strict governance protocols. This model offers the highest level of privacy and is more likely to receive participant consent, but it limits transparency, reproducibility, and the potential for secondary research. As shown in Figure~\ref{fig:graphs}, private datasets represent the majority of clinical mental health data resources today. Among text-only datasets LabWriting \cite{t_2} is kept private since it contains information about personal lives of participants. Most Audio-only \cite{a_1,a_3,a_4,a_5,a_6,a_7} and Video-only \cite{v_1,v_3,v_4,v_5,v_6} datasets are kept private since they contain biometric and identifiable features like voice and face of a person respectively. Similarly multimodal datasets \cite{ta_2,t_3,t_4,t_5,t_6,t_7,ta_4,ta_5,ta_6,ta_7,ta_8,ta_9,ta_10,ta_12,ta_14,av_2,av_6,tav_2,tav_3,tav_6,tav_7} including audio or video data is also kept private for the same reason. A few datasets with EEG \cite{t_129,t_128,t_123,t_120,t_117,t_116,t_115} are also kept private due to the risk of leaking private information.

\section*{Task Types}

Clinical mental health datasets are typically developed with specific downstream tasks in mind, reflecting key stages of diagnosis, symptom assessment, and treatment planning. Based on our analysis, these tasks fall into four primary categories: binary classification, multi-class classification, questionnaire score prediction, and therapeutic response generation. The task distribution across reviewed datasets is shown in Figure~\ref{fig:graphs}(c). Some datasets are also used for analysing the features that are present in patients with certain mental health issues and correlation of these features with severity prediction \cite{t_108,t_110,t_106,v_3,t_7,ta_5,ta_8,ta_10,ta_12,t_111,9311251,t_136,t_132,t_109}.

\paragraph{Binary Classification.} This is the most prevalent task across clinical datasets, where the goal is to determine whether an individual has a specific mental health condition, typically producing a yes/no output. These datasets are commonly used to train AI models that classify individuals as either affected by a particular disorder or as healthy controls. Most datasets try to classify depression \cite{gehrmann2018comparing,t_107,t_104,v_4,v_6,ta_1,ta_2,ta_3,t_101,av_2,av_6,av_7,av_8,tav_1,tav_2,tav_3,t_102,t_137,t_129,t_128,t_127,t_126,t_124,t_123,t_122,t_121,t_120,t_119,t_118} and schizophrenia \cite{t_2,gehrmann2018comparing,v_1,t_1,t_3,t_4,t_5,t_6,ta_4,ta_5,ta_6,ta_7,ta_9,t_103,tav_8,cobre,t_137,t_133,t_132,t_131,t_130,t_115,t_114,t_113,t_112}. Other datasets also classify anxiety \cite{mohammad-2024-worrywords,t_107,t_104,v_4,ta_3,av_2,av_7,t_102,t_137}, bipolar disorder \cite{t_107,v_5,tav_5}, and PTSD \cite{a_3,a_4,a_5,a_6,a_7,tav_2,tav_3,t_117,t_116}.
Many datasets support binary classification either directly or by design -- particularly those comparing affected individuals with healthy controls, making it well-suited for early diagnostic research. In addition to classification, these datasets are often used to analyze linguistic, acoustic, and visual markers associated with different disorders. However, binary labels offer limited clinical nuance and do not reflect symptom severity or heterogeneity.

\paragraph{Multi-Class Classification.} A smaller set of datasets support multi-class classification, which distinguishes between different severity levels, such as mild, moderate, or severe, of a given disorder \cite{a_1,a_2,v_2,ta_13,ta_14,av_5,tav_5,tav_7,t_135}. This granularity is useful for tracking disease progression or informing treatment decisions. These datasets typically include clinician-assigned severity ratings or annotated questionnaire scores that allow stratification into multiple classes. Although more clinically relevant than binary classification, this task still provides a coarse approximation of symptom complexity. Some other multi-class classification datasets also classify between several disorders. PRIORI \cite{a_1} classifies participants into suffering from depression or bipolar disorder or as healthy controls. Similarly, Aich et al. \cite{ta_13} classify between schizophrenia, bipolar disorder and healthy controls. UCLA \cite{t_134} uses MRI to classify participants into schizophrenia, bipolar disorder and healthy control classes.

\paragraph{Questionnaire Score Prediction.} Several datasets aim to predict total or item-wise scores on standardized psychological questionnaires, which are widely used in clinical settings for diagnosis and monitoring. These datasets provide rich supervision signals by aligning model outputs with instruments used in real-world practice. Table \ref{tab:ques} provides a summary of the different questionnaires used in the datasets reviewed in this paper. For depression, datasets commonly include annotations for the Hamilton Depression Rating Scale (HDRS) \cite{a_1,v_6,ta_14,av_5,t_106,t_123} and the Patient Health Questionnaire (PHQ-9) \cite{ta_3,t_110,t_111,av_2,av_7,av_8,edaic,tav_1,tav_2,modma,t_129}, with others using BDI-II \cite{v_4,av_4,t_101,t_127,t_126,t_124,t_118}, SDS \cite{ta_1,t_108,t_121}, QIDS-SR \cite{av_6}, CES-D \cite{tav_3}, HADS-D \cite{t_126} or IDS-SR \cite{t_110}. Social anxiety datasets incorporate LSAS \cite{v_2}, STAI \cite{v_4}, GAD-7 \cite{av_2,av_7,t_111,t_129}, HADS-A or SIAS and SPS \cite{t_104} , while bipolar disorder datasets use YMRS \cite{a_1,ta_14,tav_5}. PTSD-related datasets include labels based on SUD \cite{t_105}, PCL-5 \cite{a_7,tav_3} or PCL-C \cite{edaic,tav_2}. Schizophrenia datasets support prediction of PANSS scores \cite{v_1,ta_6,ta_7,tav_6,t_115}, SAPS \cite{ta_10}, SANS \cite{ta_10}, NSA-16 \cite{ta_9}, and TLC \cite{ta_10,tav_6}. These datasets are typically multimodal and annotated by clinical professionals, making them highly valuable for translational AI research. Questionnaire prediction tasks also provide more interpretable and clinically actionable outputs than raw classification.

\paragraph{Therapeutic Response Generation.} An emerging frontier in computational mental health is therapeutic response generation, where models are trained to produce contextually appropriate and empathetic responses in therapy-like settings. Among the datasets surveyed, only MEDIC \cite{tav_4} offers the necessary modality and structure to support this task. Although originally developed for empathy detection, MEDIC includes audio-visual recordings of counseling sessions, making it a valuable resource for response generation research. In contrast, existing datasets such as \cite{edaic, tav_1} primarily consist of clinical interviews aimed at diagnostic or severity assessment, lacking the dynamic and interactive nature of real therapy sessions. This highlights a critical gap: the scarcity of datasets capturing authentic therapeutic interactions. To advance the development of AI systems that can meaningfully augment therapy or assist clinicians, there is a pressing need for the collection and dissemination of high-quality, recorded therapy session datasets.

\begin{table}[!t]
\centering
\begin{tabular}{@{}lllll@{}}
\toprule
Questionnaire & Disorder& Completed By & Items & Range\\ \midrule
HDRS \cite{hamilton1960rating}& Depression & Clinician& 17& 0-52 \\
PHQ-9 \cite{kroenke2001phq}& Depression & Patient& 9 & 0-27 \\
BDI-II \cite{beck1961inventory} & Depression & Patient& 21& 0-63 \\
SDS \cite{sds}& Depression & Patient& 20& 20-80\\
QIDS-SR \cite{qids}& Depression & Patient& 9 & 0-27 \\
CES-D \cite{cesd}& Depression & Patient& 20& 0-60 \\
HADS-D \cite{hads}& Depression & Patient& 7& 0-21 \\
IDS-SR \cite{rush2000inventory}& Depression & Patient& 30& 0-84 \\
LSAS \cite{lsas} & Anxiety& \begin{tabular}[c]{@{}l@{}}Clinician/\\ Patient\end{tabular} & 24& 0-144\\
STAI \cite{stai} & Anxiety& Patient& 40& 40-160 \\
GAD-7 \cite{gad7}& Anxiety& Patient& 7 & 0-21 \\
HADS-A \cite{hads}& Anxiety & Patient& 7& 0-21 \\
SIAS \cite{sias_sps}& Anxiety& Patient& 20 & 0-80 \\
SPS \cite{sias_sps}& Anxiety& Patient& 20 & 0-80 \\
YMRS \cite{ymrs} & Bipolar Disorder & Clinician& 11& 0-60 \\
SUD \cite{sud}& PTSD & Patient& 1& 0-10 \\
PCL-5 \cite{pcl-5}& PTSD & Patient& 20& 0-80 \\
PCL-C \cite{pcl-c}& PTSD & Patient& 17& 17-85\\
SAPS \cite{saps} & Schizophrenia& Clinician& 34& 0-170\\
SANS \cite{sans} & Schizophrenia& Clinician& 25& 0-125\\
NSA-16 \cite{nsa} & Schizophrenia& Clinician& 16& 16-96\\
PANSS \cite{panss}& Schizophrenia& Clinician& 30& 30-210 \\
TLC \cite{tlc}& Schizophrenia& Clinician& 18& 0-72 \\ \bottomrule
\end{tabular}
\caption{Overview of clinical questionnaires for assessing mental health disorders, including target disorder, respondent type (clinician or self-report), number of items, and score range.}
\label{tab:ques}
\end{table}

\section*{Synthetic Datasets}

Synthetic data generation has emerged as a promising strategy to address longstanding challenges in mental health AI research, including data scarcity, privacy concerns, and limited demographic diversity. By simulating clinically realistic data, these methods enable the creation of training and evaluation resources without exposing sensitive patient information. We show a summary of existing synthetic datasets in Table~\ref{tab:synth-datasets}.

A common approach uses large language models (LLMs), such as ChatGPT \cite{openai_chatgpt_2023}, in zero- or few-shot settings to generate synthetic conversations derived from real clinical data. For example, Wu et al.\cite{ptsd-synth} extend the E-DAIC PTSD dataset\cite{edaic} using ChatGPT-generated interview transcripts. Other efforts, such as Psych8k~\cite{chat_counselor} and MentalChat16k~\cite{Mentalchat16k}, produce synthetic question-answer (QA) pairs by prompting LLMs with anonymized interview data. MDD-5k~\cite{MDD_5k} scales this further by generating thousands of synthetic diagnostic dialogues using a neuro-symbolic multi-agent LLM framework trained on real interactions. D\textsuperscript{4}\cite{D4} and Mousavi et al.\cite{mousavi-etal-2021-like} also generate simulated clinician–patient conversations using portraits or textual profiles of real individuals. While these methods reduce direct privacy risks, they still rely on real-world data and may be vulnerable to information leakage.

To address this, recent work avoids real data altogether. Datasets such as Thousand Voices of Trauma\cite{thousand_voices_of_trauma} and PSYCON\cite{psycon} construct diverse synthetic profiles of clients and therapists -- varying in age, gender, behavior, and disorder, and generate counseling dialogues across a wide spectrum of conditions, including depression, PTSD, schizophrenia, and bipolar disorder. Similarly, HealMe\cite{healme} simulates clients from the PATTERNREFRAME dataset\cite{pattern_reframe}, with dialogues generated between a ChatGPT client and therapist. However, HealMe follows a rigid, fixed-step counseling strategy, limiting conversational diversity. SMILE\cite{smile} and SoulChat\cite{soulchat} aim to increase variability by generating multi-turn dialogues from QA pairs in PsyQA~\cite{psyqa} or crowdsourced sources, though the lack of clinical grounding in these responses can limit psychological fidelity.

To address this, CBT-LLM\cite{cbt-llm} enriches QA interactions using prompts informed by cognitive behavioral therapy (CBT), while CACTUS\cite{cactus} builds on PATTERNREFRAME with more detailed client profiles and dynamic CBT-based conversations, improving both grounding and variability. CPsyCoun~\cite{cpsycoun} follows a similar pattern, using forum-sourced memos as seeds for synthetic counseling dialogues, though expert oversight remains limited.

While most synthetic datasets focus on text, some begin to address multimodal aspects of mental health. M2CoSC\cite{M2CoSC} augments textual dialogues with static client images generated via GPT-4V. However, the visual content remains unchanged during interaction, and dialogue generation follows fixed strategies. In contrast, MIRROR\cite{Mirror} simulates dynamic facial expressions in response to conversational cues and integrates a CBT-driven dialogue model, representing a significant step toward psychologically grounded, multimodal simulation. 

Despite these advances, there remains a substantial gap in synthetic generation for speech and video modalities. Non-verbal signals, such as tone, pause patterns, facial expressions, and body posture, are critical for clinical judgment and therapeutic engagement. Future research should prioritize multimodal synthetic data generation that captures these dimensions to more accurately model real-world therapeutic interactions.

\begin{table}[t]
\centering
\begin{tabular}{@{}lllll@{}}
\toprule
Dataset & Language & Modality& Size& Data types\\ \midrule
PTSD Synth \cite{ptsd-synth}& English& Text&3281(ZS)/392(FS) & Multi-turn Dialogue \\
Psych8k \cite{chat_counselor} & English& Text& 8,187 & QA Pairs\\
MentalChat16k \cite{Mentalchat16k} & English& Text& 6,338 & QA Pairs\\
CpsyCounD \cite{cpsycoun} & Chinese& Text& 3,134 & Multi-turn Dialogue \\
HealMe \cite{healme} & English& Text& 1,300 & Multi-turn Dialogue \\
SMILE \cite{smile}& Chinese& Text& 55,165& Multi-turn Dialogue \\
SoulChat \cite{soulchat} & Chinese& Text& 2,300,248 & Multi-turn Dialogue \\
CBT QA \cite{cbt-llm} & Chinese& Text& 22,327& QA Pairs\\
CACTUS\cite{cactus}& English& Text& 31,577& Multi-turn Dialogue \\
Thousand Voices of Trauma\cite{thousand_voices_of_trauma}& English& Text& 500& Multi-turn Dialogue \\
PSYCON\cite{psycon}& English& Text& 1020& Multi-turn Dialogue \\
D\textsuperscript{4}\cite{D4}& Chinese& Text& 1,339& Multi-turn Dialogue \\
Mousavi et al.\cite{mousavi-etal-2021-like}& English& Text& 800& Multi-turn Dialogue \\
MDD-5k\cite{MDD_5k}& Chinese& Text& 5,000& Multi-turn Dialogue \\
M2CoSC\cite{M2CoSC}& English& Text, Image & 429 & Multi-turn Dialogue \\
MIRROR \cite{Mirror} & English& Text, Image & 3,073 & Multi-turn Dialogue \\

\bottomrule
\end{tabular}
\caption{Overview of synthetic mental health datasets, detailing language, modalities, dataset size (ZS: zero-shot, FS: few-shot), and data type (multi-turn dialogues or single-turn QA pairs).}
\label{tab:synth-datasets}
\end{table}

\section*{Dataset Modalities}

\paragraph{Single-Modality Datasets.}  While multimodal learning offers a richer representation of mental health cues, single-modality approaches remain widely used due to their lower cost, simpler deployment, and reduced privacy risks. In some cases, disorders can be reliably detected from just one modality, making unimodal systems a viable alternative.

Text-only datasets are relatively rare and often limited in diagnostic utility due to the context-dependent nature of language. For instance, WorryWords~\cite{mohammad-2024-worrywords} links over 44,000 words to anxiety associations, but lacks conversational context. Other studies leverage clinical notes to detect disorders such as schizophrenia and depression~\cite{gehrmann2018comparing}, or analyze written language in patients with schizophrenia~\cite{t_2}. Hou et al.\cite{t_108} analyze the correlation between depression and reading habits among university students. They analyze the text of the book read by a person and try to predict if they suffer from depression. The College Health Surveillance Dataset (CHSN) \cite{t_107} contains a large amount of data from EHRs of 1 Million university students over 6 Million visits covering mental disorders like depression, anxiety and bipolar disorder. The Ex-Ray \cite{t_103} dataset also contains psychological reports for clustering patients. Several others explore text classification of schizophrenia~\cite{t_1,t_3,t_4,t_5,t_6,t_7}, though many also collect audio to analyze speech patterns.

Audio data provides valuable diagnostic signals, especially for PTSD through features like prosody, spectral characteristics, and vocal tract dynamics~\cite{a_2,a_3,a_4,a_5,a_6,a_7,t_105,t_104}, often extracted using tools like OpenSMILE~\cite{a_4,a_7} or Wav2Vec. On the other hand, PRIORI \cite{a_1} collects Smart phone conversations to predict depression and bipolar disorder through extracted rhythm features. Similarly, RADAR-MDD \cite{t_110} contains smartphone speech recordings of depressed patients from the U.K., Spain and Netherlands to find depression markers from a set of $28$ speech features. Similarly, Chang et al. \cite{t_106}  ask psychologists to analyze speech cues in smartphone recordings to understand speech features that are important for diagnosing mental illness. 
Despite its diagnostic value, audio data collection presents challenges, including equipment requirements and acoustic control.

Video introduces the greatest privacy risks due to identifiable facial features, but is particularly effective in assessing disorders involving non-verbal cues. Gaze aversion and facial expressions are key indicators of social anxiety~\cite{v_2,v_3}, depression~\cite{v_4,v_6}, bipolar disorder~\cite{v_5}, and schizophrenia~\cite{v_1}. Abbas et al.~\cite{v_1}, for instance, use head movement rates from smartphone cameras to detect schizophrenia.

Physiological modalities like EEG and MRI also show their utility in diagnosing various mental disorders. Datasets with EEG are mainly used to diagnose depression \cite{t_137,t_129,t_128,t_126,t_124,t_123,t_122,t_121,t_120,t_119,t_118} and schizophrenia \cite{t_137,t_136,t_135,t_115,t_114,t_113,t_112} but can also help in anxiety \cite{t_137} and PTSD \cite{t_117,t_116} diagnosis. Meanwhile, MRI data is useful for schizophrenia classification \cite{cobre,t_134,t_133,t_132,t_131,t_130,t_109}.

\paragraph{Multimodal Datasets.}  Many psychiatric symptoms manifest across multiple behavioral channels, making multimodal analysis essential for accurate diagnosis. For depression, relevant cues span speech semantics~\cite{chim-etal-2024-overview}, prosody~\cite{cummins2015review}, and facial expressions~\cite{slonim2023facing}. Evidence from prior works further supports the efficacy of multimodal approaches. The AVEC-2017 \cite{avec_2017} and AVEC-2019 \cite{avec_2019} challenges demonstrated that models incorporating text, audio, and visual information outperform unimodal baselines in depression severity prediction. Similarly, the AVEC-2018 challenge \cite{avec2018} showed improved performance in bipolar disorder diagnosis through multimodal fusion.

The most common combination is text and audio, seen in 22 clinical datasets targeting conditions like depression, anxiety, schizophrenia, bipolar disorder, and PTSD~\cite{ta_1,ta_2,ta_4,ta_5,ta_6,ta_7,ta_8,ta_9}. These datasets typically comprise interview transcripts and audio recordings from clinical or semi-structured interactions. Some, like MMPsy~\cite{ta_3} and Jeong et al.\cite{ta_10}, collect responses to structured prompts. Following a similar approach, Ex-Ray \cite{t_103} records participants speaking and writing a passage about a particular topic with specific target words. Other datasets embed diagnostic tasks, including Theory of Mind assessments\cite{ta_12}, social role-play~\cite{ta_13}, and narrative responses to TAT images~\cite{ta_14}. Other tasks include phoneme task, testing phonemic and semantic fluency, picture description and prompted narrating \cite{t_111}. Hayati et al. \cite{t_101} collect and transcribe semi-structured interviews for depression detection in different Malay dialects. 

Ten public datasets include all three modalities -- text, audio, and video, while seven include audio and video. To protect privacy, raw video is often withheld, and only extracted features like facial landmarks or action units are shared~\cite{av_2,av_5,av_6,av_7,edaic,tav_1,tav_2,tav_3,tav_4,tav_6,tav_7}. Most multimodal datasets are collected from interviews with trained professionals or virtual agents, though several involve task-based paradigms. AViD~\cite{av_4}, for instance, uses PowerPoint prompts to guide participants through reading, storytelling, and TAT-inspired tasks. Guo et al.\cite{av_8} balance emotional content across stimuli, while BDS\cite{tav_5} and Zhang et al.~\cite{tav_8} incorporate structured speaking tasks such as counting and passage reading. In another study, Tao et al. \cite{t_102} collect video recordings of patients talking  with a ChatGPT-based virtual character in real time mental health conversation.

Beyond behavioral data, some datasets integrate physiological or neuroimaging signals. MODMA~\cite{modma} includes paired EEG and audio for depression detection, while Zhu et al. \cite{t_127} use audio, video and EEG for depression detection. VerBIO \cite{9311251} uses EEG and audio to analyze anxiety disorder.

\begin{figure}[!t]
\centering
\includegraphics[width=\linewidth]{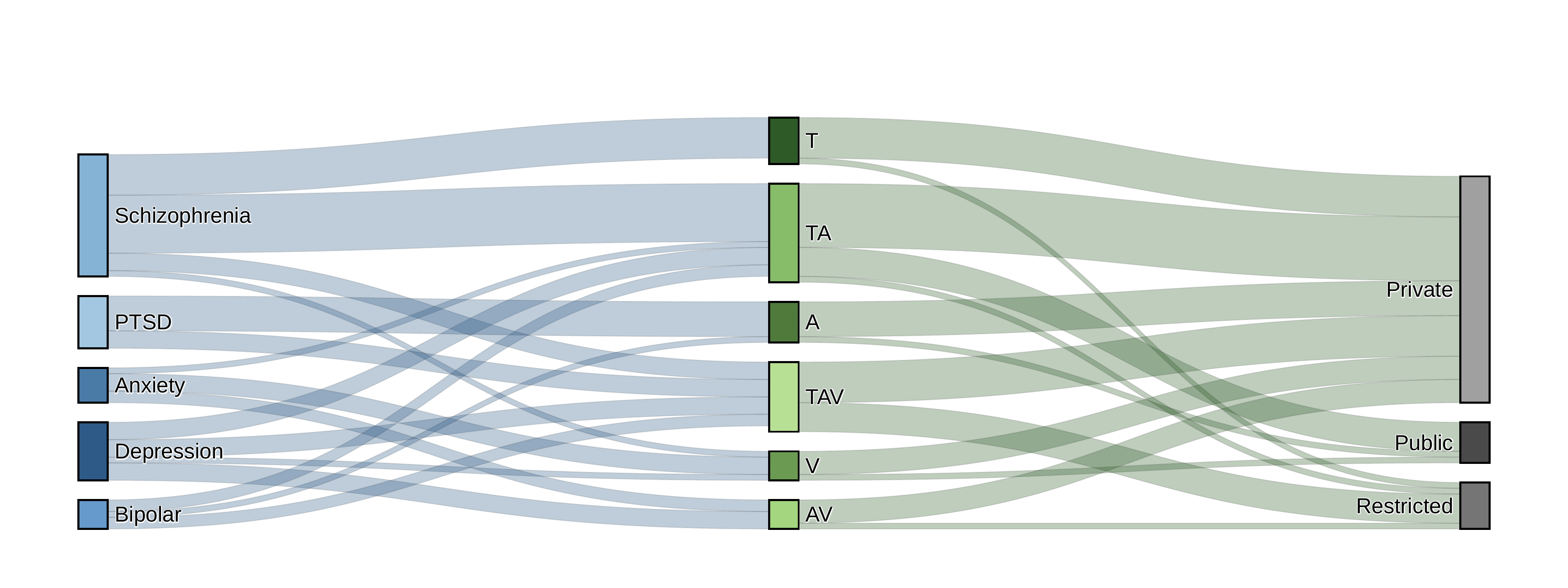}
\caption{Sankey diagram depicting the distribution of dataset modalities across mental health disorders and their accessibility levels. The figure visualizes the types and combinations of modalities present in clinical mental health datasets for various disorders, along with their accessibility status (public, restricted, or private), highlighting gaps in data availability and openness across different disorders and formats.}
\label{fig:shankey}
\end{figure}

\section*{Language and Cultural Diversity in Datasets}

Culture profoundly shapes how individuals experience, express, and interpret psychological distress~\cite{cult1}. Across cultural and linguistic contexts, people employ different metaphors for mental illness~\cite{cult2}, and their symptoms may manifest in divergent physical or behavioral forms~\cite{cult3}. Ignoring these cultural nuances risks misdiagnosis, inappropriate treatment, and reduced efficacy of AI models in mental health care.

For instance, individuals with depression in South Asian cultures often attribute their distress to supernatural or moral causes, whereas White British individuals more frequently cite biological explanations~\cite{cult4}. Cultural interpretations of anxiety also vary: Americans commonly report fear of heart attacks, while Cambodians describe sensations such as ``limb blockage'', tightness or soreness in arms and legs~\cite{cult5}. Emotional expression is likewise shaped by cultural norms. In individualistic societies such as the United States and United Kingdom, anxiety often stems from guilt or self-blame, whereas in East and Southeast Asian cultures, it is more commonly rooted in embarrassment~\cite{cult5}.

Cultural differences also influence symptomatology in psychotic and mood disorders. Although auditory hallucinations are a near-universal feature of schizophrenia, visual hallucinations are more frequently reported in Africa, Asia, and the Caribbean than in Europe or North America~\cite{cult6}, and the content of hallucinations varies by region. In bipolar disorder, Tunisian patients often present initially with manic symptoms, while French patients more commonly exhibit depressive episodes~\cite{cult7}. For PTSD, symptoms in populations such as the Kalahari Bushmen or Vietnamese communities often diverge from Western norms like emotional numbing or avoidance, resulting in underdiagnosis~\cite{cult8}.

Even LLMs, such as ChatGPT, have shown limitations in multicultural therapeutic contexts, tending by default to culturally neutral or Western-centric advice, rather than offering context-specific responses~\cite{cult9}. Zahran et al.~\cite{cult10} test eight different LLMs including multi-lingual LLMs in the Arabic context and find that current LLMs are inadequate in culture specific contexts. Hayati et al.~\cite{t_101} demonstrate that even within the same language (Malay), the performance of LLMs in depression detection can vary significantly across regional dialects, underscoring the need for cultural and linguistic sensitivity. These examples highlight the necessity for culturally competent AI systems, trained on diverse datasets that reflect the lived experiences of varied populations.

Yet, as illustrated in Figure~\ref{fig:graphs}(e), most clinical mental health datasets remain concentrated in English-speaking (e.g., U.S., U.K., Canada, Australia) and Chinese-speaking (e.g., China, Taiwan, Taipei) settings. Some additional representation exists from Germany (German), Turkey (Turkish), Pakistan (Urdu), Greece (Greek), Poland (Polish), Denmark (Danish), Italy (Italian), the Netherlands (Dutch), and Singapore (English or Mandarin), Spain (Spanish), Malaysia, Korea, Japan, Chile, Russia. However, these are typically limited to one or two datasets per region. Large global regions, including South and Southeast Asia, the Middle East, Africa, Central and South America, are still largely missing. Similarly, widely spoken languages such as Hindi, Arabic, Bengali, Portuguese, and major African languages are either severely underrepresented or absent entirely.

This lack of cultural and linguistic diversity inherently limits the generalizability of current AI models. To ensure equitable, effective, and globally relevant mental health technologies, there is an urgent need to collect and curate datasets from underrepresented cultures and languages.

\section*{Challenges}

\paragraph{Small Datasets.} Despite growing interest in clinical mental health datasets, most remain small, often under $200$ participants (Table~\ref{tab:real-datasets}), due to high collection costs, logistical challenges, and the ethical and legal complexities of handling sensitive data. Such limited scale hinders robust AI training, leading to overfitting and poor generalization. While methods like transfer learning and regularization offer partial relief, they cannot replace scale, highlighting the persistent tension between protecting privacy and enabling research.
To address these limitations, we propose federated learning and synthetic data generation. Federated learning \cite{fl} enables model training across decentralized, institution-specific datasets (Table~\ref{tab:real-datasets}) without sharing raw data, thus preserving privacy while increasing effective scale. Synthetic data, grounded in the statistical and clinical properties of real datasets, can further augment training, but must capture the nuanced complexities of mental health rather than superficial correlations.
\paragraph{Limited Diversity.} Current datasets often lack diversity in geography, language, culture, and clinical setting, being mostly collected in English- or Chinese-speaking countries and within single institutions. This narrow scope limits model generalizability and risks cultural and demographic bias. Data sharing across regions is constrained by privacy regulations, but federated learning can enable training on distributed datasets without moving sensitive data. Synthetic data generation, using diverse client and therapist profiles \cite{MDD_5k,psycon,thousand_voices_of_trauma,D4,cactus}, can further reflect varied linguistic, cultural, and demographic contexts, capturing complex contextual–psychological interactions to support more inclusive mental health technologies.
\paragraph{Lack of Standardization.} Wide variation in protocols, annotations, modalities, and task definitions hampers training and makes evaluation inconsistent. Without shared schemas, collaboration remains fragile. Addressing this requires unified annotation, metadata, and evaluation standards, alongside developing AI systems flexible enough to handle missing modalities and heterogeneous inputs via multi-task and modular architectures.
\paragraph{Synthetic Data: Missing Modalities, Languages, and Depth.} While synthetic data can address privacy concerns and expand datasets, most current resources are single-turn English or Chinese textual dialogues, omitting the multimodal, multilingual, and longitudinal nature of real psychotherapy. They rarely model temporal dynamics, underrepresented disorders, or diverse cultural contexts. Clinically useful datasets should incorporate multi-session, multimodal (audio, visual, physiological) data, multiple languages, and diverse therapist–client profiles, grounded in psychological theory and enabled by multimodal LLMs \cite{next-gpt}. Achieving this requires sustained collaboration among clinicians, computational scientists, and cultural experts. Without such interdisciplinary effort, synthetic data will remain an elegant but shallow solution to a much deeper problem.
\paragraph{Federated Learning.} Federated learning (FL) can address the scale and diversity limits of mental health datasets but remains vulnerable to gradient leakage attacks \cite{NAGY2023110475}. Combining FL with local differential privacy (LDP) \cite{NAGY2023110475} adds protection by locally injecting noise, yet often degrades performance on small clinical datasets \cite{dp-fl-benchmark}. Advancing secure, effective deployment requires novel LDP-FL methods that better balance privacy and utility, validated in realistic, resource-constrained settings.
\paragraph{Privacy.} As shown in Table~\ref{tab:real-datasets} and Figure~\ref{fig:graphs} (b), most clinical mental health datasets remain private due to a lack of robust, scalable privacy mechanisms. Traditional anonymization, often single-modality and without formal guarantees \cite{privacy-ai}, is inadequate, especially for multimodal data, where one modality can reveal sensitive attributes from another. Broader access will require multimodal anonymization methods with theoretical guarantees to prevent cross-modal leakage and ensure equitable protection across populations.

\paragraph{Evaluation Benchmark.} The value of a mental health dataset is ultimately determined by the performance of models trained on it in clinically relevant tasks. Existing efforts like CounselBench \cite{counselbench} are useful but focus mainly on single-turn counseling. There is a need for broader benchmarks covering tasks such as disorder classification, symptom severity prediction, and multi-turn therapy assessment, while reflecting cultural, linguistic, and demographic diversity. Developing such resources will require close collaboration among clinicians, computational scientists, and cultural experts to ensure both clinical validity and fairness.

\section*{Future Directions}

Advancing mental health research through AI requires a shift toward standardized, ethically sound, and privacy-preserving data collection practices. Future initiatives must adopt clear, field-wide data collection principles that not only ensure compliance with ethical and legal standards but also facilitate effective and secure use of the data for AI development. We highlight three complementary strategies for responsibly leveraging collected mental health data:
(i) Federated learning with local differential privacy (LDP-FL), where data collected across geographically distributed sites, each representing diverse cultural and demographic populations remains stored locally, with model training conducted in a privacy-preserving, decentralized manner;
(ii) Multimodal synthetic data generation grounded in psychological theory, using real-world clinical data as reference points and drawing upon diverse therapist and client profiles, to supplement existing datasets and enhance robustness of trained models;
(iii) Public release of anonymized multimodal datasets, made possible through advanced anonymization techniques with theoretical privacy guarantees to minimize the risk of re-identification while retaining research utility.

\paragraph{Data Collection.} To build large, diverse, and clinically valuable datasets, data collection must follow standardized protocols with prior approval from institutional ethics committees and informed consent from participants, clearly detailing intended data use, storage procedures, and privacy safeguards. Collection should capture the multimodal nature of therapy sessions through high-quality audio-visual recordings of patient–therapist interactions, with video documenting facial expressions, gaze, head movements, and body posture of patients, and audio recorded via separate microphones to allow accurate speaker diarization. Text transcripts should be created manually by authorized researchers or via secure, local automatic speech recognition (ASR) systems to avoid external data transmission. Additionally, physiological measures such as EEG and MRI should be included where relevant, especially for conditions like schizophrenia and depression, where they have shown diagnostic value. Adhering to these guidelines is critical for producing clinically meaningful, ethical, and reproducible datasets that support inclusive AI research in mental health.

\paragraph{Data Utilization.} To responsibly leverage mental health data while protecting privacy, we identify three complementary pathways.
First, federated learning combined with local differential privacy (LDP-FL) provides a privacy-preserving framework for using distributed datasets across institutions and regions. This approach enables integration of culturally and linguistically diverse data by keeping it locally stored. However, current LDP-FL methods often underperform on small clinical datasets \cite{dp-fl-benchmark}, highlighting the need for improved techniques balancing privacy and utility.
Alternatively, data collected within a single institution can be augmented via synthetic data generation. Multimodal large language models \cite{next-gpt} can create high-quality synthetic datasets guided by structured client and therapist profiles reflecting diverse cultural, linguistic, and demographic attributes. These synthetic interactions, should be grounded in psychological frameworks such as cognitive behavioral therapy (CBT) \cite{cactus}, and enhanced with few-shot examples from real, ethically collected data, improving realism and clinical fidelity. Crucially, synthetic data is free of real patient identifiers and can be publicly shared to promote transparency and accelerate research while preserving privacy.
Finally, public release of real-world mental health data would greatly improve research utility and remains an essential long-term goal but requires robust multimodal anonymization methods. Existing anonymization techniques are usually modality-specific and fail to address cross-modal privacy risks (e.g., audio revealing text-based sensitive information). Future frameworks must provide theoretical privacy guarantees and handle complex multimodal data. Once achieved, safe public release of diverse datasets can enable large-scale aggregation and foster more generalizable, inclusive mental health AI systems.

\section*{Author Contributions} 
T.C. and I.G. conceptualized the idea; A.M. and P.K.A. developed the work; A.M., P.K.A., T.C. and H.A. wrote the manuscript; T.C., I.G., A.M., P.K.A. and H.A. revised the manuscript.

\section*{Funding Information}
This research work has been funded by the German Federal Ministry of Research, Technology and Space and the Hessian Ministry of Higher Education, Research, Science and the Arts within their joint support of the National Research Center for Applied Cybersecurity ATHENE. This work has also been funded by the DYNAMIC center, which is funded by the LOEWE program of the Hessian Ministry of Science and Arts (Grant Number: LOEWE/1/16/519/03/09.001(0009)/98). T.C. acknowledges the travel support of the Alexander von Humboldt Foundation through a Humboldt Research Fellowship for Experienced Researchers, the support of the Rajiv Khemani Young Faculty Chair Professorship in Artificial Intelligence, and Tower Research Capital Markets for work on machine learning for social good.

\section*{Competing Interests}
The authors declare no competing interests.

\section*{Additional Information}

\noindent{\bf Materials \& Correspondence} should be emailed to Tanmoy Chakraborty (\url{tanchak@iitd.ac.in}).

\bibliography{sample}

\end{document}